%% file: ensemble_ec_optimization_gecco2020.tex
\documentclass[sigconf]{acmart}

\usepackage{booktabs} % For formal tables
\usepackage{lineno}
\usepackage{amsmath}
\usepackage{caption}
\usepackage{booktabs}
\usepackage{amsfonts}
\usepackage{algorithm}
\usepackage[noend]{algpseudocode}
\usepackage{longtable}
\usepackage{multicol}
\usepackage{rotating}
\usepackage{placeins}
 \usepackage{multirow}
\usepackage{hyperref}

\usepackage{xspace}
\usepackage{graphicx}
\usepackage{paralist}
\usepackage{url}
\usepackage{matharticle}
\usepackage{subfig}

\usepackage{xcolor}
\usepackage{colortbl}
\definecolor{GAINSBORO}{HTML}{DCDCDC}
\definecolor{LIGHTYELLOW}{HTML}{FFFFE0}
\definecolor{MISTYROSE}{HTML}{FFE4E1}
\definecolor{ALICEBLUE}{HTML}{F0F8FF}

\copyrightyear{2020}
\acmYear{2020}
\setcopyright{acmcopyright}\acmConference[GECCO '20]{Genetic and Evolutionary Computation Conference}{July 8--12, 2020}{Cancún, Mexico}
\acmBooktitle{Genetic and Evolutionary Computation Conference (GECCO '20), July 8--12, 2020, Cancún, Mexico}
\acmPrice{15.00}
\acmDOI{10.1145/3377930.3390229}
\acmISBN{978-1-4503-7128-5/20/07}

\usepackage{amsfonts,amsmath,amssymb}

\usepackage[utf8]{inputenc}
\usepackage{graphicx}
\usepackage{verbatim}

\usepackage{multirow}
\usepackage{booktabs}
\usepackage{url}
\usepackage{subfig}
\usepackage{dblfloatfix} 

\usepackage{graphicx}
\usepackage{setspace}
\usepackage{multicol}

\usepackage{xcolor}
\usepackage{colortbl}
\usepackage{amssymb}

\definecolor{GAINSBORO}{HTML}{DCDCDC}
\definecolor{LIGHTYELLOW}{HTML}{FFFFE0}
\definecolor{MISTYROSE}{HTML}{FFE4E1}
\definecolor{ALICEBLUE}{HTML}{F0F8FF}
\usepackage{pifont}% http://ctan.org/pkg/pifont

\newcommand{\REO}{\texttt{REO-GEN}\xspace}
\newcommand{\NREO}{\texttt{NREO-GEN}\xspace}
\newcommand{\REOlong}{\texttt{Restricted Ensemble Optimization of GENenrators}\xspace}
\newcommand{\NREOlong}{\texttt{Non-Restricted Ensemble Optimization of GENenrators}\xspace}
 
\newcommand{\RandomREO}{\texttt{Random REO Search}\xspace}
\newcommand{\RandomNREO}{\texttt{Random NREO Search}\xspace}
\newcommand{\RREO}{\texttt{RRS}\xspace}
\newcommand{\RNREO}{\texttt{RNRS}\xspace}

\newcommand{\iterGreedy}{\texttt{Iterative Greedy}\xspace}
\newcommand{\randomGreedy}{\texttt{Random Greedy}\xspace}

\newcommand{\RSL}{\texttt{\RREO}$\blacktriangledown$\xspace}
\newcommand{\GAL}{\texttt{GA}$\blacktriangledown$\xspace}
\newcommand{\RSNL}{\texttt{\RREO}\xspace}
\newcommand{\GANL}{\texttt{GA}\xspace}

\newcommand{\IG}{\texttt{IG}\xspace}
\newcommand{\RG}{\texttt{RG}\xspace}

\newcommand{\GA}{\texttt{GA}\xspace}

\begin{document}
\title{Re-purposing Heterogeneous Generative Ensembles with Evolutionary Computation
}

%\author{Anon Anon} 
%\affiliation{%
%  \institution{Anon, \\Anon}
%}
%\email{anon@anon.edu}
%\renewcommand{\shortauthors}{Anon et al.}

\author{Jamal Toutouh}
\affiliation{%
  \institution{Massachusetts Institute of Technology, CSAIL}
}
\email{toutouh@mit.edu}

\author{Erik Hemberg}
\affiliation{%
  \institution{Massachusetts Institute of Technology, CSAIL}
}
\email{hembergerik@csail.mit.edu}

\author{Una-May O'Reilly}
\affiliation{%
  \institution{Massachusetts Institute of Technology, CSAIL}
}\email{unamay@csail.mit.edu}

\renewcommand{\shortauthors}{J. Toutouh et al.}

\begin{abstract}
\input{abstract}
\end{abstract}




\maketitle

\input{introduction}
\input{background}

\input{ensemble-optimization}

\input{methods}
\input{experimental-setup}

\input{experimental-analysis}

\input{conclusions}

\input{ack}

\bibliographystyle{ACM-Reference-Format}
\bibliography{bibliography} 

\end{document}

%% file: abstract.tex
Generative Adversarial Networks (GANs) are popular tools for generative modeling. The dynamics of their adversarial learning give rise to convergence pathologies during training such as mode and discriminator collapse. In machine learning, ensembles of predictors demonstrate better results than a single predictor for many tasks. In this study, we apply two evolutionary algorithms (EAs) to create ensembles to re-purpose generative models, i.e., given a set of heterogeneous generators that were optimized for one objective (e.g., minimize Fr\'echet Inception Distance), create ensembles of them for optimizing a different objective (e.g., maximize the diversity of the generated samples). The first method is restricted by the exact size of the ensemble and the second method only restricts the upper bound of the ensemble size. Experimental analysis on the MNIST image benchmark demonstrates that both EA ensembles creation methods can re-purpose the models, without reducing their original functionality. The EA-based demonstrate significantly better performance compared to other heuristic-based methods. When comparing both evolutionary, the one with only an upper size bound on the ensemble size is the best.

%% file: introduction.tex
\section{Introduction}
\label{sec:introduction}

Generative Adversarial Networks (GANs) consist of a set of unsupervised machine learning methods to learn generative models~\cite{goodfellow2014generative}. 
They take a training set drawn from a specific distribution and learn to represent an estimate of that distribution. 
GANs combine a generative network (generator) and a discriminative network (discriminator) that apply adversarial learning to be trained. 
Thus, the generator learns how to create ``artificial/fake'' samples that approximate the real distribution, and
%they are able to generate ``artificial/fake'' samples that approximate the given distribution. 
the discriminator is trained to distinguish the ``natural/real'' samples from the ones produced by the generator. 
GAN training is formulated as a minmax optimization problem by the definitions of generator and discriminator loss.
Hopefully, the GAN training can converge on a generator that is able to approximate the real distribution so well that the discriminator only provides a random label for real and fake samples. 

%If we have space we could add some GAN applications

%GANs training problem
GANs have been successfully applied to a wide range of applications~\cite{wu2017survey,wang2019generative,pan2019recent}. 
However, training GANs is difficult since the adversarial dynamics may give rise to different convergence pathologies~\cite{arjovsky2017towards,arora2017gans}, e.g. vanishing gradient, gradient explosion, and mode collapse. 
When gradient pathologies appear, the generator is not able to learn and, if the problem persists, it basically generates noise for the whole training process. 
Mode(generator) collapse happens when the training converges to a local optimum, 
i.e. the generator produces realistic fake images that only represent a portion of the real data distribution. Therefore, the GAN has not successfully learned the distribution. 

A large body of work is devoted to mitigate these problems and improve GAN training robustness~\cite{ganhacks,mao2017least,gulrajani2017improved,nguyen2017dual,radford2015unsupervised}. 
A promising approach is training ensembles of networks~\cite{gulrajani2017improved,grover2018boosted,toutouh2019}. 
The ensemble is created by training multiple instances of a GAN (generator/discriminator) in different ways (e.g., different initialization seeds) and combining them (e.g., randomly drawing samples from any generator in the ensemble). 

Machine learning (ML) has been taking advantage of evolutionary computing (EC) to address many different tasks~\cite{Alexandropoulos2019,jin2006multi}. 
In addition, EC has also been applied to ensembles creation~\cite{abbass2003pareto,larcher2019autocve,islam2003constructive,langdon2002combining,liu2000evolutionary,chandra2006ensemble,zhang2017,fernandes2019}. In particular, EC and GANs have been studied with Lipizzaner~\cite{schmiedlechner2018towards,schmiedlechner2018lipizzaner} and Mustangs~\cite{toutouh2019}, which apply a distributed coevolutionary approach to train a population of GANs to create an ensemble of generators. Through the training epochs these methods use an evolutionary algorithm (EA) to learn mixture weights to optimize the quality of the generative models.  

Lipizzaner and Mustangs apply common measures to assess the the quality of a generative model, e.g.,  
Inception Score (IS)~\cite{salimans2016improved} and Fr\'echet Inception Distance (FID)~\cite{Martin2017GANs}. 
FID is widely used in image generation problems since it correlates well with the perceived quality of samples and is sensitive to dropped modes. 
However, FID is not able to distinguish between different failure cases~\cite{borji2019pros}. 
For example, generative models with competitive results in FID score may produce high quality images but only a subset of them~\cite{sajjadi2018assessing}. 

The total variation distance (TVD) is a
scalar measure used to assess the capacity of a generative model to
cover the real data distribution~\cite{li2017distributional} by
accounting for class balance.
In this study, we are concerned with the diversity in the generative
models (i.e., mode collapse). 
\sloppy
It is empirically shown that FID is not correlated with the diversity of the generated samples~\cite{borji2019pros,sajjadi2018assessing}.  
Thus, we define two problems to optimize ensembles of generators already trained in terms of FID (quality of the image generated) to obtain high diverse samples, i.e., to re-purpose already trained GANs to optimize TVD: 
\REOlong (\REO) and \NREOlong (\NREO). 
The main difference between both problems is the first one is restricted by the specific size (number of generators) of the ensemble and the second does not define the size of the ensemble but an upper bound. 
Finally, we devised two methods evolutionary algorithm (EA) methods to address them.
 
The main contributions of this paper are: 
\begin{inparaenum} [\itshape 1)]
\item proposing two optimization problems, REO-GEN and NREO-GEN, to obtain high quality and diverse samples from already trained generative models, 
\item proposing different methods to address these optimization problems,   
\item and obtaining generative models that maximize the diversity of the high quality generated samples.
\end{inparaenum}

Thus, we answer the following research questions:
\begin{asparadesc}
\item [\textbf{RQ1:}] \textit{Is it possible to re-purpose GANs trained in terms of a given objective (e.g., FID) to optimize another objective (e.g., TVD) without requiring applying the high computational cost of GAN training?}
\item [\textbf{RQ2:}] \textit{How can we create high quality and diverse ensembles?} 
\item [\textbf{RQ3:}] \textit{Can be created ensembles of GANs trained independently?}
\item [\textbf{RQ4:}] \textit{Can we use EC to create competitive ensembles?}
\item [\textbf{RQ5:}] \textit{Is there any relation between the ensemble size and the quality of the generative model created to deal with a given problem?}
  \end{asparadesc}

The paper is organized as follows. 
Section~\ref{sec:related-work} and Section~\ref{sec:tvd-gan-optimization} present the related work and the optimization problems, respectively. 
Section~\ref{sec:methods} describes the methods devised to address the problems. 
The experimental setup is in Section~\ref{sec:experimental-setup} and results in Section~\ref{sec:results}. 
Finally, conclusions are drawn and future work is outlined in Section~\ref{sec:conclusions}.

%% file: background.tex
\section{Related Work}
\label{sec:related-work}

In this section, we introduce some previous research in the main topics of our study: EC to create ensembles and GAN ensembles.  

\subsection{Evolutionary Ensemble Learning}
\label{sec:eel}

In ML, an important approach to improve the performance (e.g., classification or prediction) of the models is to strategically combine/fuse them. This is known as ensemble learning~\cite{zhang2012ensemble}. 

There is a body of work on Evolutionary Ensemble Learning (EEL), i.e., the application of EC to learn ensembles. 
The main idea is to compensate for the limitations of the models trained with the abilities of the others by combining EC with learning algorithms. 
Accurate and diverse models are the best candidates to satisfactorily create strong ensembles~\cite{thomas2000ensembles}. 
Ensemble creation includes several strategies such as~\cite{harith2019,Alexandropoulos2019}: boosting~\cite{schapire1990strength}, cascade generalization~\cite{gama2000cascade}, and stacking~\cite{wolpert1992stacked}.  Genetic Algorithms (GAs) have been used to successfully create ensembles by multiple ML methods, e.g. Support Vector Machines, Neural Networks (NN), and Genetic Programming (GP)~\cite{plawiak2019application,zhou2002ensembling,padilha2016,sohn2019}.

GP has been used as a base learner for classification or regression problems. 
Moreover, GP has been used to improve such a models by creating ensembles~\cite{bhowan2011,mauvsa2017co,folino2006gp,fitzgerald2013bootstrapping,folino2003ensemble,iba1999bagging}. 
A bagging method based on GP has been applied to ensemble Interval GP (IGP) classifiers that are independently trained (i.e., performing different runs)~\cite{tran2018genetic}. 
The experimental analyses have shown that this new method is able to outperform single IGP and different types of ensembles. 
In order to reduce the computational complexity of ruining GP training methods to obtain the models, a method based on a spatial structure with bootstrapped elitism (SS+BE) has been presented~\cite{dick2018evolving}. 
This method is able to generate an ensemble from a single run.

Multi-objective (MO) GP has been applied to build an ensemble of classifiers with unbalanced data, which competing objectives are the minority and majority class accuracy~\cite{bhowan2011}. 
Other MO approaches have been applied to generate ensembles~\cite{ribeiro2018multi,ying2019egp}. 
For example, 
Pareto Differential Evolution (PDE) has been used to evolve an ensemble of NN using negative correlation learning (NCL) in the fitness to increase ensemble diversity~\cite{abbass2003pareto}. PDE and Nondominated Sorting GA-II (NSGA-II) have been similarly applied to evolve NN ensembles. In this case, the training accuracy is traded off against the diversity of the evolved models, measured
using the NLC of the ensemble members~\cite{chandra2006ensemble}. 

There is also a lot of research about applyingapplying Evolutionary Strategies~\cite{lima2013nonlinear,schmiedlechner2018lipizzaner,toutouh2019}, Particle Swarm Optimization~\cite{pulido2016design, aburomman2016novel, ripon2020efficient} or
Ant Colony Optimization~\cite{meng2018price, ripon2020efficient}.

\subsection{GAN Ensemble Models}
\label{sec:ensemble-gan}

Recently, a number of variants of GANs have raised to address training issues suffered by the seminal approach presented by Goodfellow~\cite{goodfellow2014generative}. 
Inspired by the works that propose creating ensembles of convolutional neural networks (CNNs) as a straightforward way to improve results~\cite{krizhevsky2012imagenet,chen2019ensemble}, 
several authors have proposed similar solutions to deal with GAN learning pathologies. 
Many GAN methods have been proposed for remedying the convergence pathologies of GAN training by creating mixtures(ensembles) of GANs~\cite{gulrajani2017improved,wang2016ensembles,tolstikhin2017adagan,grover2018boosted,hoang2017multi,durugkar2016generative}. 

EC has been also applied to generate ensembles of GANs. 
Lipizzaner~\cite{schmiedlechner2018towards,schmiedlechner2018lipizzaner} and Mustangs~\cite{toutouh2019} train a spatially distributed population of GANs. 
Sub-populations (neighborhoods) of GANs apply a competitive coevolutionary algorithm to adversarially train the generators and the discriminators. 
The main motivation behind  Lipizzaner is to maintain diversity in order to improve the robustness of GAN training. 
At the end of each epoch, these algorithms apply a number of generations of the Evolutionary Strategies (ES) algorithm \cite{loshchilov2013surrogate} to evolve the mixture weights used to define the ensemble of generators defined by the sub-populations to maximize the quality of the generated samples (e.g., to optimize IS or FID). 

Thus, the main motivation of this research is to define a novel method to re-purpose (or to improve) GANs by coupling concepts coming from EEL and GANs training. 
The aim is to fill the gap that exists in addressing GAN training issues by applying a methodology motivated by the successful EEL. 
This methodology can obtain more competitive new generative models by combining heterogeneously trained generators, without requiring the high computational cost of re-training new GANs.

%% file: ensemble-optimization.tex
\section{Output Diversity GAN Optimization}
\label{sec:tvd-gan-optimization}

The goal of this study is to re-purpose generative models to allow
them to generate samples according to a new objective function. 
We focus on optimizing TVD (diversity)
of GANs that have been optimized to get high quality images in terms of FID. Eq.~\ref{eq:tvd} defines TVD applied to assess the class balance
(i.e., diversity of the samples generated), lower TVD means more
diverse samples.
\begin{equation}
\small
TVD = \frac{1}{2}\sum_{c \in Classes}{\mid freq(real_c) - freq(fake_c) \mid}
\label{eq:tvd}
\end{equation}

%In general, the likelihood of fooling the discriminator is proportional to the accuracy of the fake sample. 
In general, the quality of the generative models is assessed
by using one-dimensional metrics (e.g., IS and FID) that are not able
to measure the diversity of the
samples~\cite{sajjadi2018assessing}. We empirically evaluated the
correlation between FID and TVD scores of 700 generators trained to generate digits of MNIST
dataset~\cite{lecun1998mnist} by creating 50,000 images from each generator. 
The generators with high quality
(FID$<$40) have a Spearman's correlation coefficient between FID and
TVD of 0.2, i.e. no correlation. 
Thus, when FID is being optimized does not mean that TVD is also being improved. 
This motivates the optimization of TVD for generators that have been optimized on
FID. We have defined \REO and \NREO to create ensembles of generators that maximize
the diversity (i.e., minimize TVD) without worsening the FID.

Given a set of trained generators $\mathbf{G}=\{g_0, ..., g_{n-1}\}$.
We define the ensemble optimization problem as to find the
$n$-dimensional mixture weight vector $\vw$ that is defined as follows
\begin{equation}
\footnotesize
\vw^* =\argmin TVD\big(\sum_{0 \leq i < s} w_i G_{g_i}\big)\;,
\label{eq:mixture}
\end{equation}
where $w_i$ represents the probability that
a data sample comes from the $i$th generator in $\mathbf{G}$ and $G_{g_i}$ represents the samples created by $g_i$, with $\sum_{w_i \in \vw} w_i = 1$.
\REO applies the restriction that the ensemble size $s$ is known, and
therefore, $|w_i\neq 0|=s$ (i.e., number of mixture weights different
to 0 are $s$).  In contrast, \NREO relaxes the restriction by defining
an upper bound for the size, $|w_i\neq0|\leq s$.

%% file: methods.tex
\section{Genetic Ensemble Creation}
\label{sec:methods}

In order to address \REO and \NREO problems, we apply two evolutionary approaches based on GA. 
We use two different solutions representations to deal with the difference between these two variants of the problem (e.g., fixed size and variable size of the ensemble)
Subsequently, we also have two different types of evolutionary operators to complement their encoding.

The solutions are \textbf{randomly initialized} keeping the values in the correct range according to the representation (see sections~\ref{sec:ga-reo-encoding} and~\ref{sec:ga-nreo-encoding}), the recombination operators are versions of the \textbf{two point crossover}, and \textbf{ad hoc mutation operators} are specifically devised for these problems. 
In the following subsections, we present them, as well as the objective (fitness) function evaluation.  

\subsection{\REO Encoding and Operators.}
\label{sec:ga-reo-evolutionary}

This section describes the solution encoding and the evolutionary operators applied to address \REO by using a GA.

\subsubsection{\REO Encoding}
\label{sec:ga-reo-encoding}

A solution is represented as a vector of real numbers with two decimals of precision, 
having length $s$ (size of the ensemble). 
Thus a solution $\mathbf{e}$ is represented by \hbox{$\mathbf{e} = \langle g^0.w^0,\dots ,\ g^{s-1}.w^{s-1}\rangle$}.   
The integer part $g^i$ of these real numbers identifies the generator in the ensemble, where $g^i \neq g^j$ for $i \neq j$ 
The probability of using $g^i$ to draw a sample is given by the two digits of the fractional part ($w^i$). 
Thus, we have the following restrictions: 
having $n$ pre-trained models $g^i \in \{0, ..., n-1\}$, 
$w^i \neq 0$, and  
$\sum^{0 \leq i < s} w_i = 100\%$. 
%\TODO{Why not as a tuple $(g,w), g \in \mathbb{Z}, w \in \mathbb{R}$, then the precision is an implementation detail} 

\begin{figure}[!h]
\vspace{-0.2cm}
\includegraphics[width=0.6\linewidth]{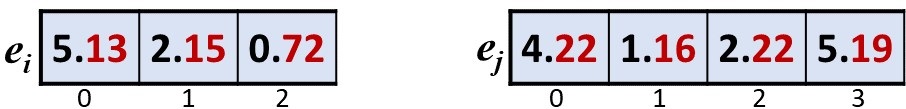}
\vspace{-0.2cm}
\caption{\REO solution representation examples.}
\label{fig:reo-representation}
\end{figure}

Figure~\ref{fig:reo-representation} solutions for two different problem instances defined by different ensemble sizes: $e_i$ and $e_j$. 
The first one represents a vector $\langle 5.13,\ 2.15,\ 0.72\rangle$ that encapsulates an ensemble of size three which is made of generators identified with numbers 5, 2, and 0 that draw samples with probabilities 0.13, 0.15, and 0.72, respectively. 
The second one shows an ensemble of size four built with generators identified with numbers 4, 1, 2, and 5 that draw samples with probabilities 0.22, 0.16, 0.22, and 0.19, respectively.

\vspace{-0.5cm}

\subsubsection{\REO Recombination}
\label{sec:reo-crossover}

\begin{figure}[!h]
\vspace{-0.2cm}
\includegraphics[width=0.65\linewidth]{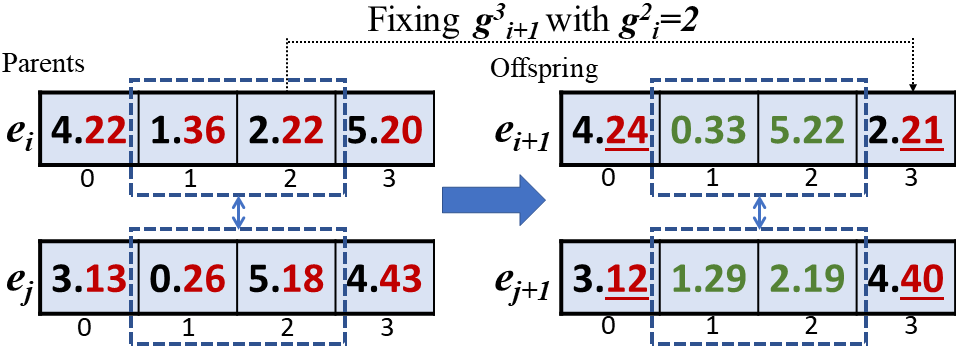}
\vspace{-0.3cm}
\caption{\REO crossover/recombination example.}
\label{fig:reo-crossover}
\end{figure}

%The recombination operator used is a version of the well-known two-point crossover. % named \REO two-point generators swap crossover. 
From the parents ($e_i$ and $e_j$), the offspring ($e_{i+1}$ and $e_{j+1}$) are generated by exchanging information of the genes (with index $c$) located between two randomly selected points within the chromosome.  
The operator swaps the integer part of the real numbers (i.e., it exchanges the generators identifier $g^c_i$ and $g^c_j$) and it averages the weights (i.e., $w^c_i$,$w^c_j$=$\frac{w^c_i+w^c_j}{2}$).    
Figure~\ref{fig:reo-crossover} shows an example of the application of this recombination operator.  

After applying the recombination a \textit{fixing operator} is applied to fix duplicate of generators and to keep the solution under restriction $\sum^{0 \leq i < s} w_i = 100$.  
It avoids duplicate generators $g_{i+1}^c$ (if there is some) by keeping the new occurrence of that generator (i.e., the one that came from the other parent) and changing the old one by the generator that was encoded in $g_{i}^c$. 
After swapping generators in Figure~\ref{fig:reo-crossover}, $g_{i+1}^2$=$g_{i+1}^3$=5, the solution is fixed by $g_{i+1}^3$=$g_{i}^2$=2. 
The weights are fixed by keeping the same proportions after computing the new weights, but normalizing them to sum 100. 
%The underlined weights have been also changed even they were not within the two points selected for the crossover. 

\vspace{-0.5cm}

\subsubsection{\REO Mutation}
\label{sec:reo-mutation}

\begin{figure}[!h]
\vspace{-0.2cm}
\includegraphics[width=0.7\linewidth]{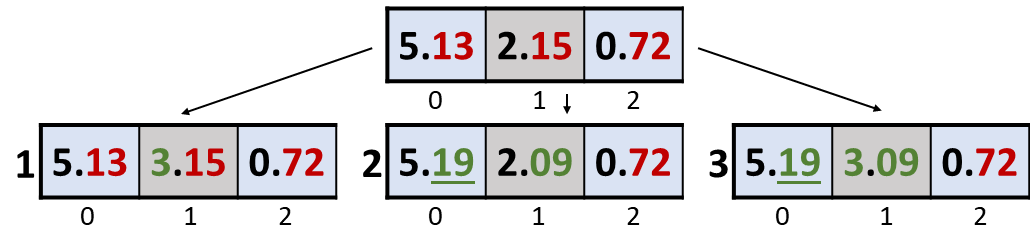}
\caption{\REO mutation examples.}
\vspace{-0.5cm}
\label{fig:reo-mutation}
\end{figure}

The mutation operator equally likely applies three variations on solutions (see Figure~\ref{fig:reo-mutation}). These variations work as follows: 
\begin{inparaenum} [\itshape 1)]
\item \textbf{changing the generator} in the mutated gene with a randomly selected one of the ones not included in the ensemble;  
\item \textbf{modifying the probabilities} of the mutated gen $w^c_i$ and another one randomly selected $w^d_i$ by adding or subtracting a random value (doing the opposite operation in the other gen) keeping the sum of probabilities in 100 and $w^c_i,w^d_i>0$; and
\item \textbf{combining both}, the generator and the weight mutations (mutation 1 and 2). 
\end{inparaenum}

%trim={<left> <lower> <right> <upper>}

\subsection{\NREO Encoding and Operators.}
\label{sec:ga-nreo-evolutionary}

This section describes the solution encoding and the evolutionary operators applied to address \NREO by applying a GA.

\subsubsection{\NREO Encoding}
\label{sec:ga-nreo-encoding}

Solutions are represented as vectors of integer numbers divided in three segments: ensemble size (one element), generators identification ($n$ elements), and probability definition ($n$ elements); 
having length $2n+1$. The ensemble 
$\mathbf{e}$ is represented by $\mathbf{e} = \langle e^0, e^1, \dots, e^n, e^{n+1}, \dots, e^{2n}\rangle$.
The first element of the vector defines the size of the ensemble, i.e, $e^0 < n$.
The following $n$ elements (generators identification segment) identify the $n$ pre-trained generators, which identifiers are randomly shuffled and located when the GA initializes the solution. 
The last $n$ elements represent the probabilities of using the generators in the same order specified in the generators identification segment, i.e., $e^{n+i}$ element of the vector encapsulates the probability of using the generator identified in the $e^i$ (i.e., $w_i$=$e^{n+i}/100$).

Thus, we have the following restrictions: 
$e^i \neq e^j$ for $i,j \in \{1, ..., n\}$ if $i\neq j$ (i.e., there is not any repetition in the generators identification segment of the vector), 
$e^i \neq 0$ $i \in \{n+1, ..., e^0\}$ (i.e., all probabilities are higher than 0), 
and $\sum_{n+1 \leq i < n+e^0} e^i = 100$. 

\begin{figure}[!h]
\includegraphics[width=0.6\linewidth]{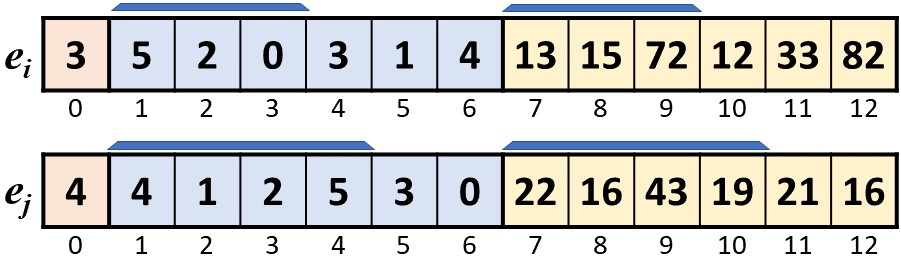}
\vspace{-0.3cm}
\caption{\NREO solution representation examples.}
\vspace{-0.3cm}
\label{fig:nreo-representation}
\end{figure}

Figure~\ref{fig:nreo-representation} illustrates two examples of solutions, $e_i$ and $e_j$, in a problem to create ensembles from six pre-trained generators. 
The first one represents an ensemble with three generators: 
generators identified with numbers 5, 2, and 0 that draw samples with probabilities 0.13, 0.15, and 0.72 of the samples, respectively. 
The second one hast four generators: 4, 1, 2, and 5, with the following probabilities: 0.22, 0.16, 0.43 and 0.19.
%The second one encapsulates an ensemble of four generators: 
%generators identified with numbers 4, 1, 2, and 5 
%that draw samples with probabilities 0.22, 0.16, 0.43 and 0.19 of the samples, respectively.   

\vspace{-0.2cm}

\subsubsection{\NREO Recombination}
\label{sec:nreo-crossover}

\begin{figure}[!h]
\includegraphics[width=0.6\linewidth]{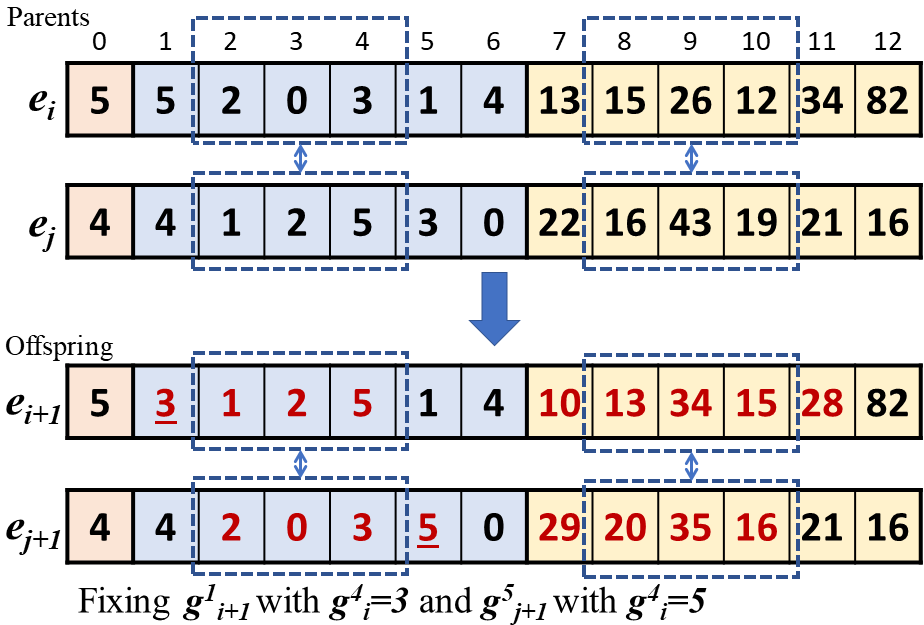}
\vspace{-0.3cm}
\caption{\NREO crossover/recombination example.}
\vspace{-0.3cm}
\label{fig:nreo-crossover}
\end{figure}

The offspring ($e_{i+1}$ and $e_{j+1}$) are generated by swapping the generators and the weights between two points in both vector segments of the parents, $e_{i}$ and $e_{j}$ (see Figure~\ref{fig:nreo-crossover}). 
The two points are selected within the range of the size of the smallest ensemble. 

Figure~\ref{fig:nreo-crossover} shows an example of applying recombination to $e_i$ and $e_j$ solutions. As the smallest ensemble ($e_j$) has size four, the two points for swapping the gens will be randomly picked between one and four. In this example, the two points are two and four. 
If after the recombination a solution has a generator duplicated, the \textit{fixing operator} is applied. 
The genetic information that has just come from the other parent is kept and
the old occurrence of that generator is changed for some generator that will be missing in the ensemble. 
The underlined generators in Figure~\ref{fig:nreo-crossover} (i.e, generator 3 in $e_i$ and 5 in $e_j$) have been modified to avoid generators repetition. 
If the sum of weights is not 100 (i.e., $\sum_{n+1 \leq i < n+e^0} e^i \neq 100$), the probability definition is fixed, too. As in \REO recombination, the weights are fixed by keeping the same proportions after computing the new weights, but normalizing them to sum 100.

\subsubsection{\NREO Mutation}
\label{sec:nreo-mutation}

\begin{figure}[!h]
\vspace{-0.3cm}
\includegraphics[width=0.57\linewidth]{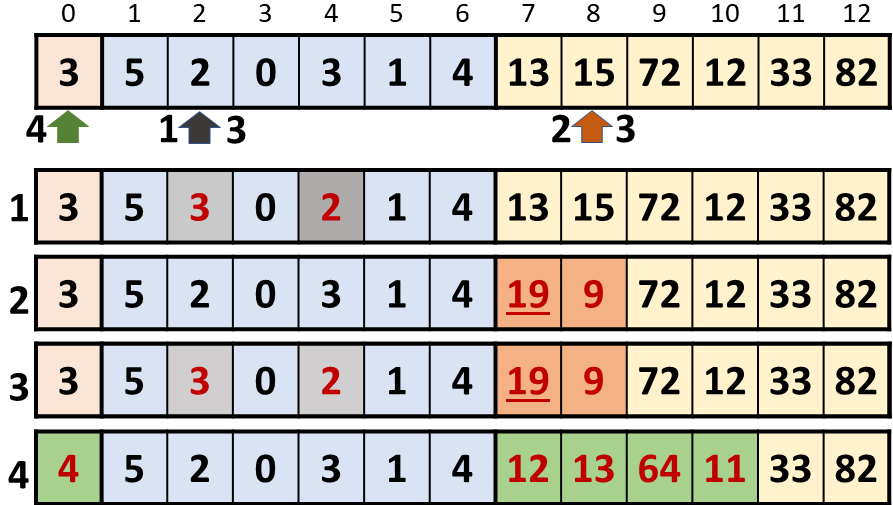}
\vspace{-0.3cm}
\caption{\NREO mutation examples.}
\vspace{-0.5cm}
\label{fig:nreo-mutation}
\end{figure}

The mutation operator equally likely applies four variations on solutions (see Figure~\ref{fig:nreo-mutation}). These variations work as follows: 
\begin{inparaenum} [\itshape 1)]
\item \textbf{exchanging a generator} in the ensemble with another one that is not included; 
\item \textbf{mutating the probabilities} by exchanging the mutated gen and another one in the probabilities definition segment (within the range of $n$ and $n$+$e^0$) by adding or subtracting a random value (doing the opposite operation in the other gen) keeping the sum of the probabilities being 100; 
\item \textbf{mutating both}, the generator and the weight; and
\item \textbf{changing the ensemble size}, which requires updating the probabilities in the way that they sum 100, but keeping proportional probabilities. 
\end{inparaenum}

%trim={<left> <lower> <right> <upper>}

\subsection{\REO and \NREO Solutions Evaluation }
\label{sec:ga-solution-evaluation}

The solutions are evaluated by drawing a number of samples (images) by using the ensemble represented in the chromosome and \textbf{assessing the resulted TVD}, i.e., \textit{fitness($e_i$)=TVD($e_i$)}.  
As stated in Section~\ref{sec:tvd-gan-optimization}, this is a \textbf{minimization problem}.

%In order to re-purpose the generative model to improve the diversity of the created samples (images) in terms of TVD (see~Eq.\ref{eq:tvd}), we draw a number of samples by using the ensemble represented by the chromosome to \textbf{evaluate the TVD}. As stated in Section~\ref{sec:tvd-gan-optimization}, this is a \textbf{minimization problem}.

%% file: experimental-setup.tex
\section{Experimental Analysis}
\label{sec:experimental-setup}

This section details the experimental analysis performed to
evaluate the evolutionary generative ensemble creators proposed to improve output diversity.

\subsection{Implementation and Execution Platform}
\label{sec:platform}

The experiments are performed on a cloud that provides 16~Intel Cascade Lake cores up to 3.8 GHz with 64~GB RAM and a GPU which are either NVIDIA Tesla P4  (8~GB RAM) or NVIDIA Tesla P100 (16~GB RAM). 
All methods have been implemented in \texttt{Python3} using \texttt{pytorch}\footnote{Pytorch Website -
  \texttt{https://pytorch.es/}} and \texttt{DEAP} (Distributed Evolutionary Algorithms in Python)~\cite{kim2019software} libraries. 
The source code implemented is public\footnote{\REO and \NREO source code -
  \texttt{https://github.com/jamaltoutouh/eel-gans}}. 

\subsection{Problem Instance -- MNIST}
\label{sec:instance}

We use the common dataset from the GAN literature: MNIST~\cite{lecun1998mnist}, which consists of low dimensional handwritten digits images.
We have had access to a pool of 700 GANs trained by using several different methods. 
After evaluating them in terms of FID and TVD, we filter out low-quality generators, we pick quality based on results of previous GAN studies based on MNIST, i.e., $FID<40$. 
Applying the threshold, we get 220~different generators with average FID of 36.393 and average TVD of 0.113. 
Finally, we address \REO and \NREO to compute ensembles that optimize TVD constructed from this set of 220 generators. 

We define different experiments for \REO and \NREO. 
First, we address the problem for ensembles sizes between three and~10 (number of classes in MNIST).  For \NREO, we
perform two different experiments, one addressing the problem with a
maximum of eight generators in the ensemble, and the second one, with
a maximum of 100 (the max ensemble size in the current implementation,
100~generators with a minimum probability of 0.01).

\subsection{Baseline Methods}
\label{sec:other-methods}

\sloppy
To evaluate the quality of the solutions from our EAs we implemented
two versions of a constructive heuristic to solve variants of the \REO
problem, named \iterGreedy (\IG) and \randomGreedy (\RG). Both methods start by sorting the generators according to the quality of the images generated (i.e, FID).  The identifier of the sorted generators are
stored on a vector $\mathbf{G^{FID}}$=$\langle g^{FID}_0, \dots,
  g^{FID}_{n-1} \rangle, g^{FID}_i \leq g^{FID}_{i+1}, i \in [0, n]$, which
the first element is the generator with the best FID.

Once the generators are sorted, \IG applies a constructive heuristic to select the generators and their weights to create an ensemble of size $s$, defined by two vectors $\mathbf{G^{FID}_o}$ that contains the generators' identifiers and $\mathbf{W^{FID}_o}$ that includes the weights (see~Algorithm~\ref{alg:ig}).%

The heuristic starts by evaluating the TVD of the current ensemble, which is defined by just one generator, i.e., $\mathbf{G^{FID}_o}=\langle g^{FID}_0 \rangle$ and $\mathbf{W^{FID}_o}= \langle 1.00 \rangle$ (line~\ref{alg:ig-evaluate-tvd}).
Then, \IG iterates over $\mathbf{G^{FID}}$ starting by the second generator to create an ensemble.

\sloppy
For each iteration $i$ a new generator $g^{FID}_i \in$ $\mathbf{G^{FID}}$ is considered to be part of the optimal ensemble, see line~\ref{alg:ig-add-new-gen} (i.e., it creates an auxiliary ensemble $\mathbf{G}$ by adding $g^{FID}_i$ to $\mathbf{G^{FID}_o}$).    
Then, \IG iteratively tests all the possible combinations of weights $\mathbf{W}$ that belong to $\mathbf{W^g}$ for the generators in $\mathbf{G}$, where $\mathbf{W^g}$=$\{0.0, 0.1, 0.2, 0.3, 0.4, 0.5, 0.6, 0.7, 0.8, 0.9\}$ to limit the possible combinations of weights and 
considering that all the combinations satisfy the restriction $\sum_{w_i \in \mathbf{W}} w_i = 1$. 
It evaluates the TVD of all the tentative generative models defined by the generators in $\mathbf{G}$ and all the combinations in $\mathbf{W}$ (line~\ref{alg:ig-evaluate-tvd2}). 
If the best tentative ensemble has better TVD than the best ensemble found so far defined by $\mathbf{G^{FID}_o}$ and $\mathbf{W^{FID}_o}$, these ensemble is considered the new best ensemble found so far. 
In contrast, if the best TVD of the new ensemble is lower, $g^{FID}_i$  is not considered to be part of $\mathbf{G^{FID}_o}$ (line~\ref{alg:ig-evaluate-criterium}). 

%It keeps the ensemble that provides the best TVD. 
At the end of each iteration $i$, if there are any generators with weight equal to 0, these generators are not considered as a part of the ensemble and they are removed from the current solution, i.e., $\mathbf{G^{FID}_o}$ (line~\ref{alg:ig-fixing}).  
This iterative method finishes when the size of the best ensemble found so far is $s$, i.e., $|\mathbf{G^{FID}_o}|=|\mathbf{W^{FID}_o}|=s$. 

\setlength{\textfloatsep}{3pt}% Remove \textfloatsep
\begin{algorithm}
  \small
  \caption{\iterGreedy \newline%: \texttt{ES}-(1+1) to evolve mixture weights $\v{w}$. \newline
		\textbf{Input:}
		~$G$: Set of generators,~$W^g$: Set of possible weights,~$s$: Size of ensemble \newline
		\textbf{Return:}
		~$\mathbf{G^{FID}_o}$: Generators in ensemble, ~$\mathbf{W^{FID}_o}$: Weights in ensemble
	}\label{alg:ig}
	\begin{algorithmic}[1] 
	 \State $\mathbf{G^{FID}_o} \gets $ sortGenerators($G$) \Comment{Sort generator by FID}
     \State $\mathbf{G^{FID}_o} \gets \langle g^{FID}_0 \rangle$ \Comment{First element of the ensemble, the best generator}
     \State $\mathbf{W^{FID}_o} \gets \langle 1.00 \rangle$ 
     \State $bestTVD \gets $ evaluateTVD($\mathbf{G^{FID}_o}, \mathbf{W^{FID}_o}$) \Comment{Evaluate TVD}\label{alg:ig-evaluate-tvd}
     \State $i \gets 1$ \Comment{Initialize counter}
	\While{sizeOf($\mathbf{G^{FID}_o}$)$<s$ \textbf{and} $i<|G|$}\Comment{Loop over generators}
     \State $\mathbf{G} \gets $ addGenerator$(\mathbf{G^{FID}_o}, g_i^{FID})$ \Comment{Add $g_i^{FID}$ to $\mathbf{G^{FID}_o}$}\label{alg:ig-add-new-gen}
	 \For{{$\mathbf{W}$} $\in$ allPossibleCombinations($\mathbf{G},W^g$)} \\ \Comment{Iterate through all possible weights}
	 \State $TVD \gets $ evaluateTVD($\mathbf{G}, \mathbf{W}$) \Comment{Evaluate TVD}\label{alg:ig-evaluate-tvd2}
	 \If {$TVD < bestTVD$}\Comment{ Replace if new ensemble is better}\label{alg:ig-evaluate-criterium}
	 \State $bestTVD \gets TVD$
	 \State $\mathbf{G^{FID}_o} \gets \mathbf{G}$
	 \State $\mathbf{W^{FID}_o} \gets \mathbf{W}$
	 \EndIf
	 \EndFor
	 \State $\mathbf{G^{FID}_o} \gets $ fixEnsemble($\mathbf{G^{FID}_o}$) \Comment{ Remove generators with weights 0}\label{alg:ig-fixing}
      %          \EndIf
     \State $i \gets i+1$
         \EndWhile
		%\State \Return $\v{w}$
	\end{algorithmic}
\end{algorithm}

\RG applies the same method, but in each iteration, it randomly picks the new generator $g^{FID}_r$ to be added to $\mathbf{G^{FID}_o}$, instead of iterating over $\mathbf{G^{FID}}$. 
The only restriction is $g^{FID}_r \notin \mathbf{G^{FID}_o}$. 
Additionally, two random search methods are applied to address both \REO and \NREO, named \RandomREO (\RREO) and \RandomNREO (\RNREO) respectively. They randomly generate a number of ensembles and return the ensemble with the best TVD.    

%\vspace{-0.5cm}
\subsection{Experimental setup}
%\vspace{-.1cm}

The solutions are evaluated by generating 50,000 samples (images) to get a robust result of the non-deterministic TVD evaluation. 
The stop condition of the \GANL and random search methods is performing 10,000 of fitness (TVD) evaluations. 
The \GANL is configured to evolve a population of 100~ensembles. 
The stop condition of the heuristic methods, \IG and \RG, is given by finding the ensemble of the required size ($s$).  
Although \IG applies an iterative deterministic method, the algorithm does not return the same result every independent run because it relies on the stochastic evaluation of the TVD. 
Moreover, this has an effect on the number of evaluations required to construct the ensemble, which may be different even for the same ensemble size. 
Thus, we perform 30~independent runs of each experiment (methods and ensemble sizes). 
To assess the statistical significance, as the results are not normally distributed, we perform Wilcoxon statistical tests (fixing p-value$<$0.01).

In addition, to compare against the heuristics for different ensemble sizes, we also report the output of \RSNL and \GANL when performing a ``similar'' number of fitness evaluations. The stop condition, in this case, is the average number of fitness evaluations performed by both greedy methods, which is different for each ensemble size. 
In the tables, these results are identified with a triangle ($\blacktriangledown$) in the name of the algorithm, i.e., \RSL and \GAL.    

Finally, owing to the stochastic nature of EAs, a parameter configuration is mandatory
prior to the experimental analysis. We performed an experimental
analysis to configure two important parameters of the \GANL: the recombination probability ($p_r$) and the mutation probability ($p_m$). For each parameter, three candidate values were
tested: $p_r \in \{0.25, 0.50, 0.75\}$ and $p_m \in \{0.01, 0.05, 0.1\}$. 
We keep the configuration that computed the best results. 
% in 10~independent runs.

%% file: experimental-analysis.tex
\vspace{-0.2cm}
\section{Results and Discussion}
\label{sec:results}
%\vspace{-0.1cm}

This section discusses the main results of the experiments carried out
to evaluate the evolutionary GAN ensembles creation problem by addressing \REO and \NREO.

\subsection{Final Fitness (TVD)}
\label{sec:final-fitness}
Tables~\ref{tbl:final-fitness-reo} and~\ref{tbl:final-fitness-nreo}
summarize the results by showing the best fitness (TVD) for the different
methods and ensemble sizes in the 30~independent runs for \REO and
\NREO, respectively.  These results are not normally distributed.
Thus, the tables present the best solution found (Min), the median
value (Median), the interquartile range (Iqr), and the least
competitive result (Max).

\begin{table}[h!]
\renewcommand{\arraystretch}{0.9} % General space between rows (1 standard)
  \small
  \centering
  \caption{\REO final fitness (TVD). Best value is \textbf{bold}.}
    \vspace{-0.3cm}
  \label{tbl:final-fitness-reo}
  \begin{tabular}{lrrrr}
      \hline
    \rowcolor{GAINSBORO}
    Method & Min (Best) & Median & Iqr & Max \\
    \hline
\rowcolor{ALICEBLUE}
\multicolumn{5}{c}{Ensemble size ($s$): 3} \\
\hline
\IG	&	0.067 &  0.074  &  0.005 &  0.086\\ %30	greddy-iterative-3
\RG	&	0.055 &  0.069  &  0.014 &  0.089\\ %30	greddy-random-3
\RSL	&	0.048 &  0.064  &  0.014 &  0.074\\ %30	optimize-random-search-3-100
\GAL	&	0.053 &  0.062  &  0.006 &  0.071\\ %30	optimize-ga-3-100
\RSNL	&	0.036 &  \textbf{0.046}  &  0.004 &  0.051\\ %30	optimize-random-search-3-10000
\GANL	&	\textbf{0.030} &  \textbf{0.046}  &  0.011 &  0.061\\ %30	optimize-ga-3-10000
      \hline
\rowcolor{ALICEBLUE}
\multicolumn{5}{c}{Ensemble size ($s$): 4} \\
\hline
\IG	&	0.056 &  0.066  &  0.005 &  0.073\\ %30	greddy-iterative-4
\RG	&	0.045 &  0.065  &  0.010 &  0.079\\ %30	greddy-random-4
\RSL	&	0.047 &  0.060  &  0.008 &  0.067\\ %30	optimize-random-search-4-530
\GAL	&	0.034 &  0.052  &  0.008 &  0.062\\ %30	optimize-ga-4-530
\RSNL	&	0.038 &  0.046  &  0.004 &  0.052\\ %30	optimize-random-search-4-10000
\GANL	&	\textbf{0.030} &  \textbf{0.039}  &  0.005 &  0.050\\ %30	optimize-ga-4-10000
      \hline
\rowcolor{ALICEBLUE}
\multicolumn{5}{c}{Ensemble size ($s$): 5} \\
\hline
\IG	&	0.047 &  0.063  &  0.007 &  0.071\\ %30	greddy-iterative-5
\RG	&	0.044 &  0.063  &  0.018 &  0.096\\ %30	greddy-random-5
\RSL	&	0.040 &  0.053  &  0.006 &  0.067\\ %30	optimize-random-search-5-1984
\GAL	&	0.034 &  0.046  &  0.005 &  0.062\\ %30	optimize-ga-5-1984
\RSNL	&	0.036 &  0.047  &  0.007 &  0.064\\ %30	optimize-random-search-5-10000
\GANL	&	\textbf{0.033} &  \textbf{0.042}  &  0.008 &  0.056\\ %30	optimize-ga-5-10000
      \hline
\rowcolor{ALICEBLUE}
\multicolumn{5}{c}{Ensemble size ($s$): 6} \\
\hline
\IG	&	0.054 &  0.060  &  0.007 &  0.075\\ %30	greddy-iterative-6
\RG	&	0.040 &  0.050  &  0.012 &  0.063\\ %30	greddy-random-6
\RSL	&	0.042 &  0.049  &  0.004 &  0.053\\ %30	optimize-random-search-6-6122
\GAL	&	0.036 &  0.044  &  0.007 &  0.055\\ %30	optimize-ga-6-6122
\RSNL	&	0.042 &  0.048  &  0.003 &  0.053\\ %30	optimize-random-search-6-10000
\GANL	&	\textbf{0.033} &  \textbf{0.043}  &  0.008 &  0.055\\ %30	optimize-ga-6-10000
      \hline
\rowcolor{ALICEBLUE}
\multicolumn{5}{c}{Ensemble size ($s$): 7} \\
\hline
\IG	&	0.046 &  0.057  &  0.003 &  0.070\\ %30	greddy-iterative-7
\RG	&	\textbf{0.030} &  0.050  &  0.007 &  0.062\\ %30	greddy-random-7
\RSNL	&	0.035 &  0.047  &  0.006 &  0.052\\ %30	optimize-random-search-7-10000
\GANL	&	0.035 &  \textbf{0.043}  &  0.005 &  0.050\\ %30	optimize-ga-7-10000
      \hline
\rowcolor{ALICEBLUE}
\multicolumn{5}{c}{Ensemble size ($s$): 8} \\
\hline
\RSNL	&	\textbf{0.036} &  0.047  &  0.006 &  0.053\\ %30	optimize-random-search-8-10000
\GANL	&	\textbf{0.036} &  \textbf{0.044}  &  0.006 &  0.050\\ %30	optimize-ga-8-10000
\rowcolor{ALICEBLUE}
\multicolumn{5}{c}{Ensemble size ($s$): 9} \\
\hline
\RSNL	&	0.034 &  0.047  &  0.004 &  0.053\\ %30	optimize-random-search-9-10000
\GANL	&	\textbf{0.031} &  \textbf{0.044}  &  0.007 &  0.054\\ %30	optimize-ga-9-10000
      \hline
\rowcolor{ALICEBLUE}
\multicolumn{5}{c}{Ensemble size ($s$): 10} \\
\hline
\RSNL	&	0.042 &  0.048  &  0.004 &  0.054\\ %30	optimize-random-search-10-10000
\GANL	&	\textbf{0.037} &  \textbf{0.044}  &  0.005 &  0.053\\ %30	optimize-ga-10-10000
      \hline
  \end{tabular}
\end{table}

Focusing on \REO results in Table~\ref{tbl:final-fitness-reo}, we observe that for the ensemble sizes 8, 9, and 10, there are no results from the heuristic methods because of computational cost restrictions. 
Additionally, for size 7, there are no results of \GAL and \RSL because the heuristics on average performed 17,630~fitness evaluations ($>>$10,000).  
\GANL obtained the best results in terms of the median value for all ensemble sizes. 
Wilcoxon statistical analysis confirms that \GANL is the best method for all ensemble sizes. 

When evaluating the second most competitive algorithm, there are two cases: first, \GAL is reported (ensemble sizes from 3 to 6), and second, there is not \GAL (ensemble sizes from 7 to 10).  
In the first case, \RSNL gets the second-best results for small ensembles sizes (3 and 4), i.e, a very limited number of fitness evaluations, and \GAL is the second-best for ensemble sizes of 5 and 6.  As \GAL performs more generations, it converges to better solutions. 
Thus, the evolutionary operators guide \GANL to more competitive results than \RSNL requiring a lower computational cost.  
In the second case, \RSNL keeps being the second most competitive. 

Regarding the best ensemble found (minimum TVD), similar results are obtained:  
\GANL is the most competitive except for ensemble size 7. 
\RG finds the best ensemble so far that size (TVD=0.030), but it required an average of 17,630~fitness evaluations. 

Regarding the methods with similar computational cost (i.e., \IG, \RG, \RSL, and \GAL), \GAL obtains the best median value and the best minimum (except for the smallest ensemble). 
%It is noticeable that the best results when comparing among the different ensemble sizes, the best results are obtained by the ensembles with size 4 (see Table~\ref{tbl:final-fitness-reo}).  
%We discuss this results in Section~\ref{sec:general-quality}.  

Focusing \NREO results in Table~\ref{tbl:final-fitness-nreo}, we can clearly state that \GANL significantly outperforms \RSNL in both types of experiments. 
These results are confirmed by the Wilcoxon statistical test (i.e., p-value$<<$0.01).
\GANL solving the second variant of the problem provides significantly better results than when addressing \REO.

\begin{table}[h!]
  \vspace{-0.2cm}
\renewcommand{\arraystretch}{0.9} % General space between rows (1 standard)
  \centering
    \small
  \caption{\NREO final fitness (TVD). Best value is \textbf{bold}.}
  \vspace{-0.2cm}
  \label{tbl:final-fitness-nreo}
  \begin{tabular}{lrrrr}
          \hline
\rowcolor{GAINSBORO}
    Method & Min (Best) & Median & Iqr & Max \\
    \hline
\rowcolor{ALICEBLUE}
\multicolumn{5}{c}{Maximum ensemble size: 8} \\
\hline
\RSNL	&	0.045 &  0.050  &  0.004 &  0.056\\ %30	new-optimize-random-search-8
\GANL	&	\textbf{0.025} &  \textbf{0.034}  &  0.007 &  0.042\\ %30	new-optimize-ga-8
\rowcolor{ALICEBLUE}
\multicolumn{5}{c}{Maximum ensemble size: 100} \\
\hline
\RSNL 	&	0.048 &  0.052  &  0.003 &  0.059\\ %30	new-optimize-random-search-200
\GANL	 &	\textbf{0.026} &  \textbf{0.035}  &  0.006 &  0.042\\ %30	new-optimize-ga-200
      \hline
  \end{tabular}
  \vspace{-0.2cm}
\end{table}
The results obtained by using \GA to address \REO and \NREO allow us to answer
\textbf{RQ4:} \textit{Can we use EC to create better ensembles?} 
Yes, EC can be used to create better generative models that are able to be re-purposed to optimize a new objective (TVD). 

\subsection{Fitness (TVD) Evolution}
\label{sec:tvd-evolution}

\begin{figure}[!h]
\vspace{-0.1cm}
\includegraphics[width=0.79\linewidth]{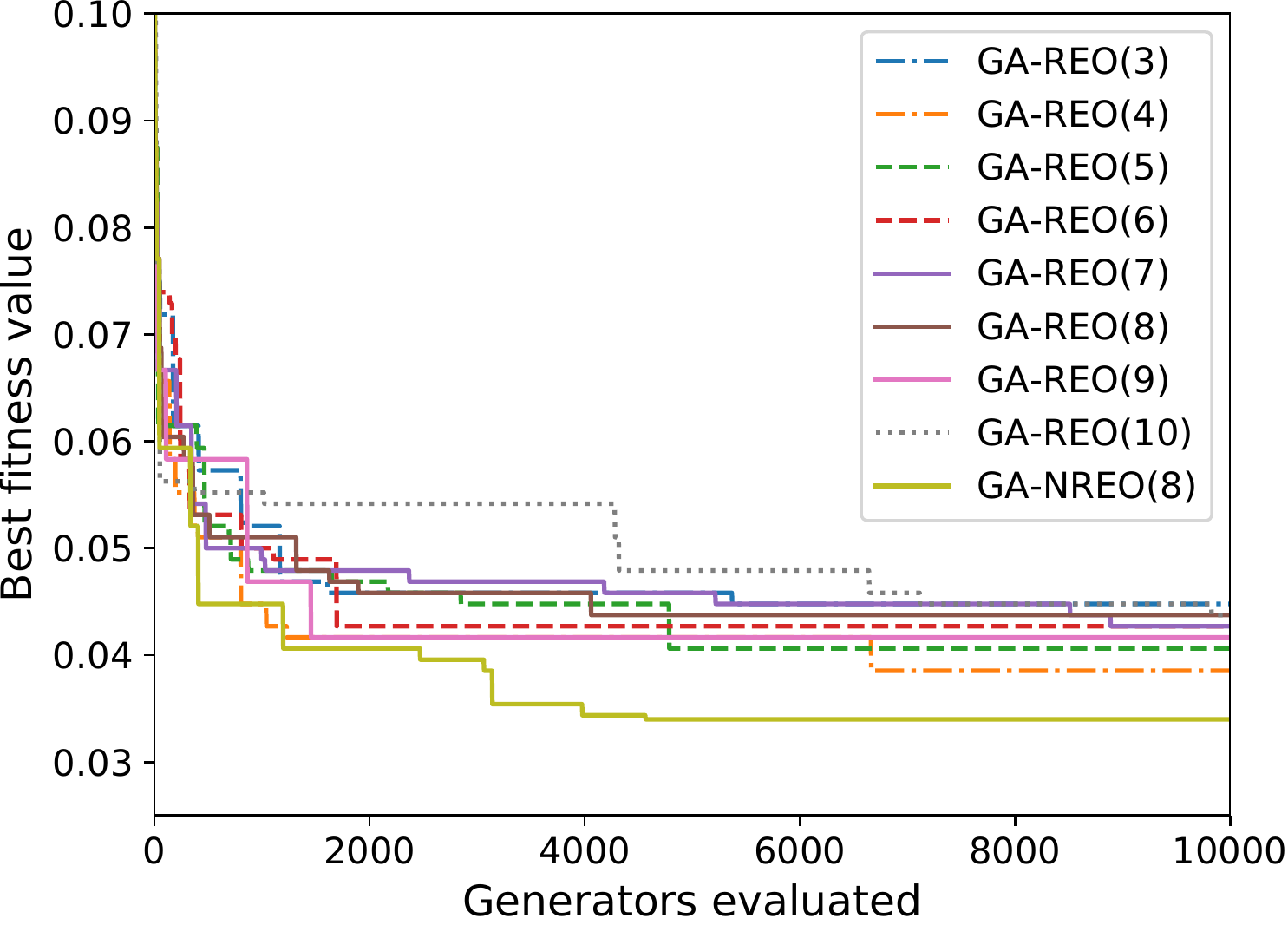}
\vspace{-0.2cm}
\caption{Fitness evolution of the median run.}
\label{fig:fitness-evolution}
\vspace{-0.2cm}
\end{figure}

Figure~\ref{fig:fitness-evolution} shows the evolution of fitness during the independent run that returns the median fitness. 
We evaluated that for \GA addressing \REO (for all values of $s$) and \NREO, which in the figure are named \texttt{GA-REO(s)}, \texttt{GA-NREO(8)}, respectively. 

We can see in Figure~\ref{fig:fitness-evolution} that \texttt{GA-NREO(8)} converges faster than the other methods to ensembles with low TVD.
It converges to the best ensemble before the first 5,000 fitness evaluations. 
When analyzing the fitness evolution of the \GA addressing \REO, when $s$=10 the algorithm converges slower and it converges to the least competitive results (to the same results than ensembles of size 3). For the other ensemble sizes, the performance is similar. However, for the ensemble size $s$=4, the fitness converges to the best results of the \texttt{GA-REO(s)}. 
This is consistent with the results in Table~\ref{tbl:final-fitness-reo}, where the best results are provided by \GANL for $s$=4.

\subsection{Comparison between Optimizers}
\label{sec:optimizers}
In order to compare the different methods, 
we evaluate the generative models returned in each independent run by \GANL and \RSNL when solving \REO (for all values of $s$) and the ones computed by \GANL when addressing \NREO (with a maximum size of 8).
Thus, we compute the percentage of improvement in the TVD of method $A$ over method $B$ by evaluating $\Delta(A, B)$=$\frac{TVD(B)-TVD(A)}{TVD(B)}\%$ for the same ensemble size. 
%\GA(\NREO) is evaluated is always for the maximum size of 8 experiment (\texttt{GA-NREO(8)}). 
Additionally, we performed Wilcoxon statistical tests to assess statistical significance. 
Table~\ref{tbl:tvd-delta} presents the results of evaluating $\Delta$. A diamond $\diamond$ is used to specify that there is statistical significance (i.e., $\alpha<0.01$).   

\begin{table}[h!]
\setlength{\tabcolsep}{4pt}
\renewcommand{\arraystretch}{0.9}
  \centering
    \small
\vspace{-0.1cm}
  \caption{Delta between methods in different ensemble sizes.}
  \vspace{-0.3cm}
  \label{tbl:tvd-delta}
  \begin{tabular}{lrrr}
      \hline
\multirow{1}{*}{Ensemble} & \texttt{GA-REO(s)} \textit{vs}& \texttt{GA-NREO(8)} \textit{vs}& \texttt{GA-NREO(8)} \textit{vs}\\ 
size ($s$) & \texttt{RRS-REO(s)}  & \texttt{RRS-REO(s)} & \texttt{GA-REO(s)} \\ 
    \hline

 3 &  -1.12\% & $\diamond$ 37.81\% & $\diamond$ 39.37\% \\ 
 4 & $\diamond$ 14.42\% & $\diamond$ 37.19\% & $\diamond$ 19.90\% \\ 
 5 & $\diamond$ 14.56\% & $\diamond$ 46.30\% & $\diamond$ 27.71\% \\ 
 6 & $\diamond$ 9.62\% & $\diamond$ 42.50\% & $\diamond$ 30.00\% \\ 
 7 & $\diamond$ 7.64\% & $\diamond$ 39.37\% & $\diamond$ 29.48\% \\ 
 8 & $\diamond$ 5.75\% & $\diamond$ 37.92\% & $\diamond$ 30.42\% \\ 
 9 & $\diamond$ 8.20\% & $\diamond$ 41.98\% & $\diamond$ 31.23\% \\ 
 10 & $\diamond$ 7.81\% & $\diamond$ 45.21\% & $\diamond$ 34.69\% \\
     \hline
  \end{tabular}
  \vspace{-0.1cm}
\end{table} 

As it can be seen in Table~\ref{tbl:tvd-delta}, when comparing \texttt{GA-REO(s)} and \texttt{RRS-REO(s)}. 
The first one is always statistically better than \texttt{RRS-REO(s)} for all ensemble sizes, except for the smallest ensemble ($s$=3) in which \GA lightly worsen the results of the random search. As shown in Figure~\ref{fig:fitness-evolution}, for this ensemble size the \GA converges to the least competitive fitness values. 
The best improvements of \texttt{GA-REO(s)} over \texttt{RRS-REO(s)} are obtained for ensembles of size 4 and 5. 
The improvements of \texttt{GA-NREO(8)} over \texttt{RRS-REO(s)} that are shown in the third column of Table~\ref{tbl:tvd-delta} are very significant ($>$37\%). 
The same occurs when we compare the ensembles obtained by this evolutionary approach and the \GA solving \REO. The ones got by the first one are statistically better for all ensemble sizes.

With these results we can answer
\textbf{RQ2:} \textit{How can we create high quality and diverse ensembles?}
The best way to obtain a generative model based on ensembles to create the digits of MNIST is by solving \NREO by using \GA. 
The grade of freedom of \GA when solving \NREO allows the method to explore better the search space to find diverse and better (high quality) solutions.

\subsection{Generative Models Quality}
\label{sec:general-quality}

In this section, we evaluate the performance of the generative models obtained in terms of TVD and FID. But, before we evaluate the shape (number of generators) of the generative models returned by \GA when it addresses \NREO since we observe that the best generator size for solving \REO is $s$=4 (see Section~\ref{sec:final-fitness}). 
Table~\ref{tbl:ensemble-shapes} shows the percentage of solutions given the size of the ensemble.

\begin{table}[h!]
\setlength{\tabcolsep}{4pt}
\renewcommand{\arraystretch}{0.9}
  \centering
    \small
      %\vspace{-0.1cm}
  \caption{Delta between methods in different ensemble sizes.}
  \label{tbl:ensemble-shapes}
    %\vspace{-0.3cm}
  \begin{tabular}{l|rrrrrrrr}
      \hline
Ensemble size & 3 & 4 & 5 & 6 & 7 & 8\\ 
\hline
Percentage of solutions ($\%$)& 23.3 & \textbf{46.7} & 20.0 & 6.7 & 0.0 & 3.3\\ 
    \hline
  \end{tabular}
    %\vspace{-0.1cm}
\end{table} 

According to the results in Table~\ref{tbl:ensemble-shapes}, the most repeated size of a final solution given by \GA is 4. 
Thus, this method likely converges to ensembles with the size of the most competitive results addressing \REO, but obtaining more competitive TVD.

Table~\ref{tbl:tvdvsfid-ensemblesize} presents the mean (Mean) and standard deviation (Stdev) of the TVD and FID obtained by the ensembles returned by \RSNL and \GANL for each value of $s$. 
Moreover, we report the same metrics for the ensembles computed by \GA addressing \NREO (last row).  
The results of individually evaluating the 220 generators used to define the problem instance are shown in the first row.

%Now, we want to evaluate the accuracy/quality of all the computed generative models according to the ensemble size. 
%Thus, we evaluate all the ensembles returned by \RSNL and \GANL for each value of $s$ in terms of TVD and FID. 
%Besides, we add the ensembles computed by \GA addressing \NREO. 
%Table~\ref{tbl:tvdvsfid-ensemblesize} presents the mean (Mean) and standard deviation (Stdev) of the evaluated metrics. 
%The first row of the table shows the results of evaluating the 220 generators used to define the problem instance.  

\begin{table}[h!]
  \centering
    \small
\renewcommand{\arraystretch}{0.9}
  \vspace{-0.1cm}
  \caption{Quality of samples according to the ensemble size.}
  \label{tbl:tvdvsfid-ensemblesize}
    \vspace{-0.3cm}
  \begin{tabular}{lrr}
      \hline
    \multirow{2}{*}{Ensemble size ($s$)} & \multicolumn{1}{c}{TVD} & \multicolumn{1}{c}{FID} \\
  & Mean$\pm$Stdev & Mean$\pm$Stdev \\
    \hline
     \rowcolor{GAINSBORO}
1	&	0.113$\pm$0.010 &	36.393$\pm$1.985 \\
3	&	0.046$\pm$0.006 &	27.576$\pm$3.947 \\
4	&	\textbf{0.043$\pm$0.005} &	27.890$\pm$3.048 \\
5	&	0.046$\pm$0.007 &	28.225$\pm$3.888 \\
6	&	0.045$\pm$0.005 &	27.077$\pm$3.030 \\
7	&	0.045$\pm$0.005 &	27.556$\pm$3.531 \\
8	&	0.046$\pm$0.005 &	26.726$\pm$3.515 \\
9	&	0.046$\pm$0.005 &	26.919$\pm$3.181 \\
10	&	0.047$\pm$0.004 &	27.218$\pm$3.408 \\
    \hline
\texttt{GA-NREO(8)}	&	 0.033$\pm$0.004 &	27.342$\pm$3.530\\
    \hline
  \end{tabular}
    \vspace{-0.1cm}
\end{table}

The use of the evolutionary created generative models (ensembles) critically improves the output produced in terms of TVD and FID, when comparing them with single generators results. 

The TVD of the samples generated by solving \NREO is significantly the most competitive one.  
Focusing on the ensembles created by addressing \REO, the best mean results are obtained by the ensembles of size 4. 
The Wilcoxon statistical test confirms that the TVD of the samples generated by the ensembles of size 4 is the most competitive with p-value$<<$0.01 (i.e., there are statistical differences between these ensembles and the others). 
The competitive results are produced by the generative models with 8 to 10 generators. 
The FID keeps similar good values between 26.7 ($s$=8) and 28.2 ($s$)=5. 
There are not any statistical differences between the ensemble sizes. 
%This is mainly due to the quality of the images generated by the selected (220) generators that are already good since they are selected because of this metric (i.e., FID$<$40). 

The results in Table~\ref{tbl:tvdvsfid-ensemblesize} allows us to answer
\textbf{RQ3:} \textit{Can be created ensembles of networks trained independently?}   
Yes, we can since the results of the ensembles critically improved the results of the networks trained independently. Thus, these networks are able to work together to provide better TVD (and even better FID), no matter how they were optimized. 

The complete set of results until now give us the information to answer
\textbf{RQ5:} \textit{Is there any relation between the ensemble size and the quality of the generative model created to deal with a given problem?}
Yes, there is a relation between the ensemble size and the quality of the generative model created. 
The results and the statistical analysis show that the ensembles that produce higher quality and diversity MNIST samples with the generators of our instance are the ones with size 4. 
Actually, the worst results have been obtained by the ensembles of sizes 3 (too small) and 10 (too big).

\subsection{Generators in Ensembles}
\label{sec:generators-in-ensembles}

\begin{figure}[!h]
\vspace{-0.1cm}
\includegraphics[width=0.4\linewidth]{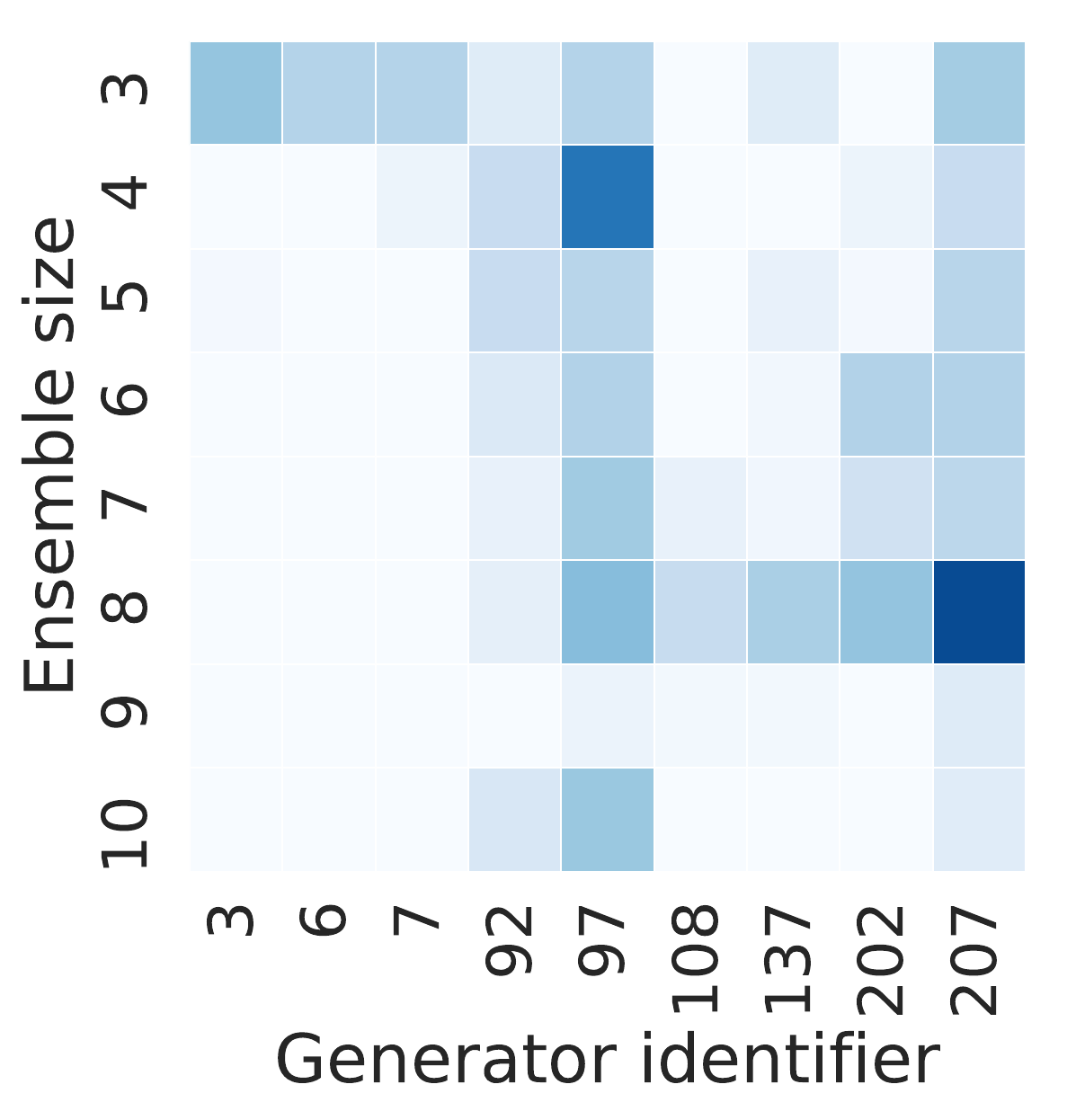}
\vspace{-0.1cm}
\caption{Frequencies of the most used generators.}\label{fig:generators-in-ensemble}
\vspace{-0.1cm}
\end{figure}

Here, we want to investigate if there is a set of predominant generators in the built ensembles 
(i.e., those that figure more often in the returned solutions). 
Thus, we measure the frequency of the generators in the returned solutions for all the computed generative models grouped by ensemble size. 
Figure~\ref{fig:generators-in-ensemble} illustrates the frequencies of the ensembles that appear at least in more than 2\% of the solutions. 
Darker blue indicates the more frequently used generator in the ensemble construction.  

The smallest size of generators ($s$=3) includes several generators with the smallest identifiers (3, 6, and 7). 
This is mainly because the \IG constructed most of its solutions with these generators, \IG iterates through the generators starting from the one with index 0. 
But it also includes generator number 207.

Focusing on the most used generators for all ensembles, 
there are three strong generators (that appear in most of the ensemble sizes): generator 92, 97, and 207. 
These generators probably are able to generate diverse types of MNIST digits.

\subsection{Computational Cost}
\label{sec:computational-cost}

An important advantage of re-purposing the GANs is we get generative models without requiring the computational cost of training the GANs from the beginning to optimize the new objective.  
Here, we report the average computational time of running the different methods presented here. 

The heuristic greedy methods (\IG and \RG) spent between 2.1~minutes (mins) for $s$=3 and 307.2~mins (when $s$=7).  
This last high computational cost makes these methods impractical to find ensembles with larger ensembles. 

The \GA spent 181.2~mins on average to address \REO and 204.8~mins on average to finish when solving \NREO. 
In the case of the proposed EAs, there is not a correlation between $s$ and the computation time, since the stop condition is performing 10,000 fitness evaluations.  

It seems considerable computational cost, but these evolutionary methods are able to return a population of 100 competitive generative models. 
The average computational cost spent to train each one of the GAN used to create the ensembles is about $~$60~mins. 

It is noticeable that according to the evolution of the fitness shown in Figure~\ref{fig:fitness-evolution}, both \GA proposed here are able to converge to ensembles with high output diversity within the first 5,000~fitness evaluations. 
Therefore, they could be executed for half of the time and they still could keep returning very competitive results.

Thus, we can see the benefit of applying the proposed methodology and answer to
\textbf{RQ1:} \textit{Is it possible to re-purpose GANs trained in terms of a given objective (e.g., FID) to optimize another objective (e.g., TVD) without requiring applying the high computational cost of GAN training?} 
is yes. Since the \GANL re-purposes the generative models requiring a bounded computational time, which is much lower than re-train a number of models to optimize them for the new objective (if it is possible to optimize such a new objective).

%% file: conclusions.tex
\section{Conclusions and Future Work}
\label{sec:conclusions}

We have empirically shown that EEL can be applied to re-purpose GANs by optimizing the diversity, as measured by TVD, of the generated samples. 
Specifically, 
starting from a number of generators previously trained, 
we defined an optimization problem to combine them in ensembles able to enhance the diversity of the samples generated. 
We devised two heuristics (greedy), \IG and \RG, and two different evolutionary approaches to address the problem. 
The first evolutionary approach addressed \REO, in which the size of the ensemble is specified as a restriction. The second evolutionary approach solved \NREO, which restriction is the maximum size of the ensemble.

These methods were tested on the MNIST. 
The results demonstrated that both evolutionary approaches construct generative models that greatly improve the diversity of the pre-trained generators while requiring a bounded computational cost. 
The best TVD results were obtained when the \GANL was used to address \NREO. 
The degree of freedom of searching for ensembles without specific sizes allows this method to move through the search space in the way that it converges to better results than the other methods.    

Even the greedy methods were the least competitive ones, they also found ensembles that improved diversity. 
However, they were able to find small ensembles since they become impractical due to the required computational cost when the sizes increased.

When evaluating the quality of the computed ensembles, we have shown that they critically improve the TVD, but they also improve the FID. 
Additionally, ensembles of size four provided the best results for the problem of generating MNIST digits. 
Actually, 46.7\% of the ensembles returned by the \GA that solves \NREO had this specific size. 

Future work will include 
the evaluation of other evolutionary approaches that will encode the weights in a different way such as a larger set of integers or continuous values. 
We will dive into the predominant generators in the built ensembles to find specific characteristics. 
In addition, we will address the problem applying a multi-objective approach in order to see the capability of such a methodology in finding ensembles that simultaneously optimize quality (FID) and diversity (TVD). 
Moreover, we want to evaluate our evolutionary approach with other datasets that require more complex networks (e.g, CIFAR-10, CIFAR-100 or CelebA).  
%Finally, this study will be a first step of devising an algorithm that will train a population of GANs (keeping them as diverse as possible) and then it will apply one of the defined methods to get high competitive generative models. 

%% file: ack.tex
\section*{Acknowledgments}
\label{sec:acks}
This research was partially funded by European Union’s Horizon 2020 research and innovation programme under the Marie Skłodowska-Curie grant agreement No 799078 and by H2020-ICT-2019-3, RTC-2017-6714-5, TIN2017-88213-R, and UMA18-FEDERJA-003 projects. 
The Systems that learn initiative at MIT CSAIL.